\documentclass[letterpaper]{article} % DO NOT CHANGE THIS
\usepackage{aaai24}  % DO NOT CHANGE THIS
\usepackage{times}  % DO NOT CHANGE THIS
\usepackage{helvet}  % DO NOT CHANGE THIS
\usepackage{courier}  % DO NOT CHANGE THIS
\usepackage[hyphens]{url}  % DO NOT CHANGE THIS
\usepackage{graphicx} % DO NOT CHANGE THIS
\usepackage{natbib}  % DO NOT CHANGE THIS AND DO NOT ADD ANY OPTIONS TO IT
\usepackage{caption} % DO NOT CHANGE THIS AND DO NOT ADD ANY OPTIONS TO IT
\usepackage{algorithm}
\usepackage{algorithmic}

\usepackage{newfloat}
\usepackage{listings}
\DeclareCaptionStyle{ruled}{labelfont=normalfont,labelsep=colon,strut=off} % DO NOT CHANGE THIS
\floatstyle{ruled}
\newfloat{listing}{tb}{lst}{}
\floatname{listing}{Listing}
\usepackage{amsfonts}
\usepackage{amsmath}
\usepackage{commath}
\usepackage{bm}
\usepackage{mathtools}
\usepackage{subcaption}
\usepackage{makecell}
\newcommand{\N}{\mathcal{N}}
\newcommand{\E}{\mathbb{E}}
\newcommand{\I}{\mathbb{I}}
\DeclareMathOperator*{\argmax}{arg\;max}
\newcommand{\Mean}{{\mathbb{E}}}

\title{Effect Size Estimation for Duration Recommendation in Online Experiments: Leveraging Hierarchical Models and Objective Utility Approaches}
\author{
    Yu Liu\equalcontrib,
    Runzhe Wan\equalcontrib, 
    James McQueen, 
    Doug Hains, 
    Jinxiang Gu, 
    Rui Song
}
\affiliations{
    Amazon.com Inc\\
    \{liuyu0jlu, runzhe.wan\}@gmail.com, 
    \{jmcq, dhains, gjinxian, ruisong\}@amazon.com
}

\begin{document}

\maketitle

\begin{abstract}
The selection of the assumed effect size (AES) critically determines the duration of an experiment, and hence its accuracy and efficiency. 
Traditionally, experimenters determine AES based on domain knowledge. 
However, this method becomes impractical for online experimentation services managing numerous experiments, and  a more automated approach is hence of great  demand. 
We initiate the study of data-driven AES selection for online experimentation services by introducing two  solutions. 
The first employs a three-layer Gaussian Mixture Model considering the heteroskedasticity across experiments, 
and it seeks to estimate the true expected effect size among  positive experiments. 
The second method, grounded in utility theory, aims to determine the optimal effect size by striking a balance between the experiment's cost and the precision of decision-making. 
Through comparisons with baseline methods using both simulated and real data, we showcase the superior performance of the proposed approaches. 
% Connecting this problem to utility theory opens the door to leverage additional optimization techniques as future work. 
\end{abstract}

\section{Introduction}
Sample size determination (SSD) plays a pivotal role in online experiments, answering the critical question of "how long should an experiment run?"~\cite{richardson2022bayesian}. 
In the spheres of online A/B testing and related classic application such as clinical trials, it is paramount to ascertain the minimal sample size during the experiment planning phase. 
If the sample size is too small, it may not accurately represent the whole population being studied, thereby introducing both bias in the results that limit drawing conclusions or generalizing the inferences. 
It can also compromise the statistical power of the experiment, leading to inaccuracies in detecting meaningful effects~\cite{ramsey2012statistical,lenth2001some}. 
Conversely, if the sample size is selected as too large, it could unnecessarily extend the duration of the experiment and inflate the associated costs, such as the human and hardware resources and  opportunity costs~\cite{wan2023experimentation}. 
Therefore, selecting an appropriate sample size is imperative to maintain a balance between decision accuracy and resource utilization.

Among the numerous methodologies for SSD in the literature ~\cite{kelley2006sample, adcock1997sample,  lindley1997choice, weiss1997bayesian},  one crucial step is to specify the assumed effect size (the absolute or percent lift from the treatment over the control). 
For example, it is a critical component in statistical power analysis used for SSD, in either the frequentist or the Bayesian setting ~\cite{du2016bayesian}. 
First of all, we need to distinguish the \textit{true effect size} of an experiment and the \textit{assumed effect size (AES)}. 
The first is an objective unknown quantity, while the latter is a subjective manually specified number. In traditional power analysis problems (e.g., in clinical trials), AES  mainly reflects (i) the experimenter's expectation (e.g., what level of improvement would be regarded as acceptable) and (ii) the trade-off between the opportunity cost of running a longer experiment and the accuracy. AES does not necessarily relate to the true effect size (or its estimate). Moreover, using observed power based on the true effect size of an experiment to determine its sample size is inappropriate, due to its 1-1 mapping to p-value~\cite{hoenig2001abuse}. 
For example, in case where an experiment lacks a noticeable effect size, the resulting observed power tends to be small. 
Using this value in SSD can lead to overestimating the required experiment duration.

    Traditionally, the AES is established by domain experts or experimenters, rather than by the experimentation services itself. For instance, \citeauthor{lenth2001some} (\citeyear{lenth2001some}) offers general guidance for selecting appropriate AES. However, within the sphere of online experiments - a setting where tens of thousands of experiments are conducted annually - a significant number of experimenters may lack the requisite domain knowledge or statistical acumen to define an appropriate AES. Additionally, statistical consultants are often resource-constrained, limiting their ability to offer personalized guidance to each team.
    In such scenarios, harnessing the vast amount of data gleaned from similar past experiments can be invaluable in determining the AES. Employing data-centric methodologies, such as meta-analysis, can automatically direct experimenters towards an appropriate AES for power analysis. This empowers them to make decisions grounded in robust, empirical evidence.

	 \textbf{Contribution.} 
    To the best of our knowledge, this paper is the \textit{first} work on data-driven effect size recommendation for large-scale online experiments in the literature. 
    Our contributions are three-fold: 
    \begin{itemize}
        \item We design a three-layer Gaussian mixture model for the distribution of observed effect sizes across experiments, which considers experiment heteroskedasticity and identifies positive ones. 
        We develop an EM-type algorithm for parameter estimation; 
        \item We also propose a utility-maximization approach to determine the optimal AES. This optimization of utility function aims to achieve a balance between experiment cost and accuracy; 
        \item We run the first large-scale empirical analysis for effect size recommendation in the literature, using real data from Amazon. 
	The analysis clearly demonstrates the effectiveness of the proposed methods. 
    \end{itemize}
    % We propose two novel approaches: 
	% (1) ; (2) 

\section{Related Work}\label{sec:related_work}
Our work is closely related to the SSD problem, for which numerous methodologies have been studied in the literature. For example, one approach involves selecting a sample size to achieve a narrow confidence interval for the standardized mean difference~\cite{kelley2006sample}. Alternatively, sample sizes can be determined based on utility theory~\cite{adcock1997sample,  lindley1997choice} or to ensure the Bayes factor exceeds a predetermined value~\cite{weiss1997bayesian}. 
However, in all these works, the effect size is assumed as a hyper-parameter that has been pre-specified by experimenters, which is not practical for online experimentation services. 
% particularly in frequentist settings. Further, researchers extend the concept of Frequentist power to Bayesian contexts, accounting for the uncertainty of the assumed effect size. This leads to a 'hybrid' power approach for Bayesian SSD~\cite{du2016bayesian}.

To our knowledge, the only work related to AES estimation is~\citeauthor{du2016bayesian} (\citeyear{du2016bayesian}). 
The authors use a random-effect meta-analysis model to quantify the uncertainty of power analysis due to the uncertainty over the effect size. 
However, the main focus is still on SSD, and no systematic study on AES recommendation is done. 
% Moreover, practical application shows that most online experiments lack a discernible effect size~\cite{wan2023experimentation}. Consequently, using this approach often leads to an underestimation of the AES.

\section{Preliminaries and Notations}
\label{section:preliminaries}

Online A/B experiments are widely used to compare customer responses between the existing feature A and the new feature B, guiding online companies' feature launch decisions. To reduce the noise and established the causal relationship in online experiments, it is standard to track customers' outcomes (e.g. ad clicks) after a customer is triggered based on some event (e.g., displaying ads).
	
Let the duration of the experiment be defined in terms of a certain number of weeks, which helps mitigate the influence of daily and weekend fluctuations. To maintain the simplicity, we assume the experiments have the same duration $t$ and omit the time subscript $t$ in notations.
Denote $n_{g}$ as the total number of customers who are triggered up to week $t$ in group $g \in \{T, C\}$ (T for the treatment and C for control). Denote sets of those
customers who are triggered up to week $t$ as $I_{g}$. Denote $Y_{i}$ as the observed outcome for customer $i$ up to week $t$, and $\bar{Y}_{g} = \sum_{i \in I_{g}} \frac{1}{n_g}Y_{i}$ is the sample mean of responses for customers in group $g$. Assuming that $Y_i$ are independent and identically distributed (i.i.d.) with mean $u_g$ and finite variance $\sigma^2_g$, as $n_. \rightarrow \infty$, we can apply the central limit theorem (CLT)~\cite{casella2021statistical}:
\begin{equation*}
    \bar{Y}_{g} | \mu_g \sim \N (\mu_g, \frac{\sigma^2_g}{n_g}), g\in \{T, C\}.
\end{equation*}
Define the average treatment effect (ATE) $\mu = \mu_T - \mu_C$, often estimated by the sample mean difference. Taking into account the independence among different customers, the sample mean difference follows as 
\begin{equation}
\label{eq:mean_diff}
    \bar{Y}_{T} -  \bar{Y}_{C}| \mu_T, \mu_C \sim \N (\mu, \sigma^2)
\end{equation}
where $\mu = \mu_T - \mu_C$ and $\sigma^2 = Var(\bar{Y}_{T} - \bar{Y}_{C}) =\frac{\sigma^2_T}{n_T} + \frac{\sigma^2_C}{n_C}$. 

%Under Bayesian setting, we further assume
%\begin{equation*}
%\mu \sim N(\mu_0, \tau^2)
%\end{equation*}
%with known $\tau^2$ and $\mu_0$.

Assume all $\sigma_g^2$ are known, the Null hypothesis under Frequentist setting and its alternative for one-sided test are
\begin{eqnarray*}
    H_0:   \mu \leq 0   \quad H_1:  \mu > 0. 
\end{eqnarray*}
When $n_. \rightarrow \infty$, \textit{frequentist power} is computed as:
\begin{eqnarray}
\label{eq:power}
	\text{power}(\delta, \sigma_., n_., \alpha ) 
	&\approx&
\Phi( z_{\alpha}   + \frac{\delta}{\sigma} ), 
\end{eqnarray} 	
where $\alpha$ is Type I error. $\delta$ is AES. $z_.$ represents the standard normal quantile. $\Phi(.)$ stands for the cumulative distribution function of the standard normal distribution. To achieve a desired power of $b$ for the alternative hypothesis, considering for a given effect size $\delta > 0$ and an allocation rate of $n_T/n_C = p$, the minimum sample size $n_T, n_C$ are selected to satisfying $power(\delta, \sigma_., n_., \alpha ) \geq b$. The required minimum duration can be determined through sample size prediction~\cite{richardson2022bayesian}.
\\

%\textbf{Standardized effect sizes}, such as Cohen's d~\cite{cohen2013statistical}, are often used in meta-analysis across different experiments, serving to normalize variations in ATE magnitudes. In this paper, a specific standardization methods isn't obligatory. However, we do require that the estimator of (standardized) effect size follows an asymptotic normal distribution. 
Standardized effect sizes, such as Cohen's d~\cite{cohen2013statistical}, are often used in meta-analysis across different experiments, serving to standardize variations in ATE magnitudes. If the estimators of standardized effect size follow an asymptotic normal distribution, all models presented in this paper can seamlessly accommodate the use of standardized effect sizes. Consequently, a dedicated discussion on this topic is omitted in this paper.

\subsection{Pooled Effect Size}
\label{section:pooled_effect_size}
\textbf{Problem Setup.} Consider $m$ past similar experiments, where $\delta_i$ is the true effect size for experiment $i$, and the observed effect size $d_i$ serves as the estimator for the true effect size. $d_i$ is commonly derived from the observed customer outcomes, e.g. Equation~\eqref{eq:mean_diff}.  We assume that observed effect sizes $\{d_i\}_{i= 1}^m$ are independent. 

The following model is used to estimate the AES~\cite{du2016bayesian}:
\begin{eqnarray} 
\label{eq:base_line}
\nonumber
	\delta_i &\sim& \N (\mu_0, \tau^2)	 \\
		\nonumber
	d_i & = &  \delta_i + e_i \\
	e_i &\sim& \N\bigg(0,  \sigma_i^2 \bigg), 
\end{eqnarray}
where we further assume  $\{\delta_i\}_{i= 1}^m$ and $\{e_i\}_{i= 1}^m$ are mutually independent.
  $\sigma_i^2$ varies across different experiments due to \textit{heteroscedasticity}, evident in Equation~\eqref{eq:mean_diff}, 
  where the variance of mean difference depends on $\sigma^2_g$ and $n_g$ from each experiment. 
  $\sigma_i^2$ can be estimated by $\hat{\sigma}_i^2 = \frac{\hat{\sigma}^2_{T, i}}{n_{T, i}} + \frac{\hat{\sigma}^2_{C, i}}{n_{C, i}}$. The estimated AES is the Maximum
  Likelihood Estimation (MLE) estimator of $\mu_0$, which is a weighted linear combination of $d_i$, thus it was also referred to as the pooled effect size.

In an online company setting, not every experiment yields a measurable effect size. This stems from the nature of online experiments with their short cycles, which was designed to encourage the exploration of innovative ideas. However, the drawback is that a significant number of ideas may not achieve statistical significance. We
categorize experiments into three groups: 1) true positive, 2) true negative, and 3) flat experiments which have no significant effect. Using $d_i$ from all experiments to train the model~\eqref{eq:base_line} will lead to an underestimation of AES, consequently elongating the required duration. The guideline is to train the model~\eqref{eq:base_line} using experiments where we know there is an underlying true positive effect. While it might seem intuitive to use hypothesis testing to categorize experiments into significant positive, negative or flat ones, the existence of Type I and Type II errors poses a challenge in precisely determining which experiments truly have such effects.

\section{Three-Layer Heteroscedastic GMM}
\label{section:fre}

\begin{figure*}[t]
  \centering
  \begin{subfigure}{.3\textwidth}
    \includegraphics[width=\linewidth]{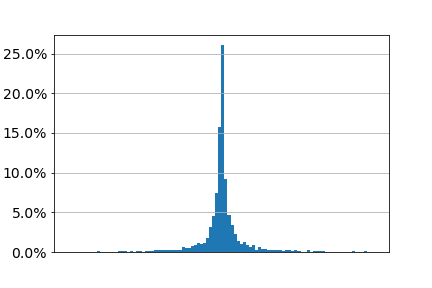}
  \caption{Histogram of observed effect sizes for certain outcome of interest among 3,300 real experiments. }
    \label{fig:subfig1}
  \end{subfigure}
  \hfill
  \begin{subfigure}{.3\textwidth}
    \includegraphics[width=\linewidth]{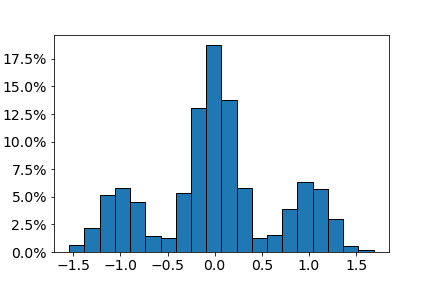}
    \caption{The simulated \textit{true} effect sizes, with three clusters corresponding to positive, flat, and negative. }
    \label{fig:subfig2}
  \end{subfigure}
  \hfill
  \begin{subfigure}{.3\textwidth}
    \includegraphics[width=\linewidth]{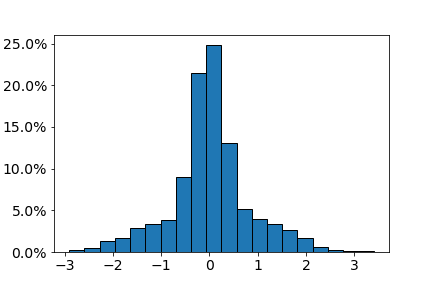}
    \caption{The simulated \textit{observed} effect sizes, with three clusters corresponding to positive, flat, and negative.}
    \label{fig:subfig3}
  \end{subfigure}
  \caption{Illustration of the motivation of using GMM. The x-axis is the effect size and y-axis is the frequency. Data in (a) are from real experiments. The x-axis
in this plot is not annotated owing to business
confidentiality. (b) was simulated through a two-layer Gaussian Mixture Model (GMM) with mean values of $(-1, 0, 1)$, variances of $(0.2^2, 0.2^2, 0.2^2)$, and component weights of $(0.2, 0.6, 0.2)$. (c) was simulated from the same model as (b) with variances of $(0.7^2, 0.3^2, 0.7^2)$.}
  \label{fig:hist_GCCP}
\end{figure*}

%In section~\ref{section:pooled_effect_size}
In the previous section, we highlighted that the main challenge of the state-of-the-art method lies in the accurately identifying experiments with true positive effect. It is natural to consider the categorization of experiments as latent variables and use the Gaussian Mixture Model (GMM).

\textbf{Motivation. }
We first illustrate our motivation of using GMM in Figure \ref{fig:hist_GCCP}. 
Figure \ref{fig:subfig1} presents the distribution of the \textit{observed} effect sizes  for certain outcome of interest over 3,300 experiment run within Amazon. 
Although at the first glance, this graph seems to support a normal distribution, we argue that it indeed illustrates the challenge of this problem. 
In Figure \ref{fig:subfig2}, we simulate the \textit{true} effect sizes following a two-Layer GMM, and in Figure \ref{fig:subfig3} we increases random errors to each of them to generate the \textit{observed} effect size, which looks like a single mode Gaussian distribution. 
Therefore, under the assumption that there exists latent clusters of positive, flat and negative experiments, it is actually infeasible to identify those positive ones with simple rules;
%as in Section~\ref{section:preliminaries}
 instead, a more principled statistical model as GMM should be used. The two-layer GMM didn't account for the required heterogeneity in our setting. %(Section~\ref{section:preliminaries}). 
 Thus, we propose a three-layer heteroscedastic GMM.

\textbf{Model.}
Assume $K=3$ is the number of clusters corresponds to clusters containing the negative, flat and positive experiments, and $m$ is the total number of past experiments. 
Without loss of generality, we assume the means of these clusters are decreasing from $K$ to $1$. 
Therefore, $k=2$ represents the cluster of flat experiments. 

Latent variable $\bm{z} = \{z_1, z_2, \ldots, z_m\}$, $z_i \in \{1, \ldots,  K \}$  represent the index class sampled from the categorical distribution parameterized by $\bm{\pi}=(\pi_1, \ldots, \pi_K)$. 
$\mu_k, \tau^2_k$ and $\pi_k$  correspond to the mean, variance, and weight of the $k$-th Gaussian component. Denote $\bm{\theta} = \{\mu_1, \ldots,  \mu_K, \tau_1,  \ldots,  \tau_K \}$.
We have 
	\begin{eqnarray} 
	\nonumber
		z_i &\sim & Categorical (k, \bm{\pi})\\
			\nonumber
		\delta_i | z_i = j &\sim& \N(\mu_j, \tau^2_j) \\
		\label{eq:three-layer-model}
		d_i| \delta_i &\sim& \N(\delta_i, \sigma_i^2), 
	\end{eqnarray}
where, the third layer arises from the heteroskedasticity among different experiments. Note that the model above is equivalent to:
%(see  Appendix $B.1$) 
		\begin{eqnarray} 
		\nonumber
		z_i  &\sim & Categorical (k, \bm{\pi})\\
		d_i | z_i = j &\sim& \N(\mu_j , \tau_j^2 + \sigma_i^2),
		\label{eq:gmm-model}
	\end{eqnarray}
where $\sigma^2_i$ are known and can be estimated using observed outcomes of the experiment $i$, e.g. Equation~\eqref{eq:mean_diff}. We can further pre-specify the means of k = 2 cluster as zero, i.e. $\mu_2 = 0$.

%Note that the marginal density is 	$p(d_i) = \sum_{k = 1}^3 p(z_k)p(d_i | z_k) =  \sum_{k = 1}^3 \pi_k N(d_i | \mu_k, \sigma^2_k)$. 

\textbf{EM algorithm.} We propose the following EM algorithm to estimate the unknown parameters in the model~\eqref{eq:gmm-model}.We first derive a few formulas that are essential for developing the EM algorithm. The marginal distribution is 
	\begin{eqnarray*}
f(d_i | \bm{\theta}, \bm{\pi})
&= &\sum_{k = 1}^ K   \pi_k f_k(d_i| \bm{\theta}), 
\end{eqnarray*}
where $f_k(d_i| \bm{\theta})  $ is the density of Gaussian distribution $\N(\mu_k, \tau_k^2 + \sigma_i^2)$. The conditional probability function for $z_i $ given the data $d_i$ is:
	\begin{eqnarray*}
	k_i(z_i| d_i,\bm{\theta}, \bm{\pi}) &=&  \frac{  \prod_{k=1}^K \bigg( \pi_k f_k(d_i|  \bm{\theta}) \bigg)^{\I[z_i=k]} }{\sum_{k = 1}^ K  \pi_k f_k(d_i|  \bm{\theta}) }. 
\end{eqnarray*}

$\I[.]$ represents the indicator function. Due to the independence among experiments, the joint distribution is 
	\begin{eqnarray*}
p(\bm{d}, \bm{z}| \bm{\theta}, \bm{\pi} )
&=&  \prod_{i = 1}^{m} f(d_i| \bm{\theta}, \bm{\pi}) k_i(z_i| d_i,\bm{\theta}, \bm{\pi}) \\
&=&  \prod_{i = 1}^{m}\prod_{k=1}^K \bigg( \pi_k f_k(d_i| \bm{\theta}) \bigg)^{\I[z_i=k]}, 
\end{eqnarray*}
where $\bm{d} = (d_1, \ldots, d_m)$.

 \textbf{E-step.} calculates the conditionally expected log-likelihood, where the expectation is conditioned on the parameters from the previous iteration step $p$ $(\bm{\theta}^{(p)},  \bm{\pi}^{(p)})$ and the expectation is taken over the latent assignments. 
\begin{eqnarray*}
\nonumber
&&Q(\bm{\theta},  \bm{\pi} |\bm{\theta}^{(p)},  \bm{\pi}^{(p)} )\\
\nonumber
&=& \sum_{i = 1}^m \E_{z|d} \bigg[log\bigg(  p_i(d_i, z_i |\bm{\theta},  \bm{\pi}  )  \bigg) |\bm{\theta}^{(p)},  \bm{\pi}^{(p)} \bigg]\\
\nonumber
&=& \sum_{i = 1}^m  \sum_{k= 1}^K \E_{z|d} \bigg[  {\I[z_i=k]}|\bm{\theta}^{(p)},  \bm{\pi}^{(p)} \bigg]\\
&&\cdot \bigg[ log( \pi_k) +  log(f_k(d_i |\bm{\theta}))   \bigg] 
\label{em:q-function}
\end{eqnarray*}

Denote the posterior probability of each Gaussian mixture component $k$ given each observation $d_i$ as:
 \begin{eqnarray}
 \label{eq:e-step}
 \nonumber
     \omega_{i,k}^{(p)} &=&\E_{z|d} \bigg[  {\I[z_i=k]}|\bm{\theta}^{(p)},  \bm{\pi}^{(p)} \bigg]\\
     &=& \frac{\pi^{(p)}_k f_k(d_i|  \bm{\theta^{(p)}}) }{\sum_{j = 1}^ K  \pi^{(p)}_j f_j(d_i|  \bm{\theta^{(p)}}) }
 \end{eqnarray}

 % (details can be found in Appendix B.2)
  \textbf{M-step.} involves finding the optimal values for $\bm{\theta},  \bm{\pi}$ by maximizing the log-likelihood derived in the E-step. The parameters in the $(p+1)$-th iteration are updated as follows: 
 \begin{eqnarray}
 \pi_j = \frac{1}{m} \sum_{i=1}^m  \omega_{i,j}^{(p)},  \label{eq:weight1}
\end{eqnarray}
where $\hat{\mu}_j$ and $\hat{\tau}^{2}_j$  are solved simultaneously by: 
\begin{eqnarray}
 &&\mu_j = \Big[\sum_{i=1}^m\frac{ \omega_{i, j}^{(p)} d_i }{(\sigma^2_i+\tau_j^{2} )}   \Big] / \Big[\sum_{i=1}^m\frac{ \omega_{i, j}^{(p)} }{(\sigma^2_i+ \tau_j^{2} )}\Big] \label{eq:mu1} \\
  % \small
 &&\sum_{i = 1}^{m}   \frac{\omega_{i, j}^{(p)}}{\sigma_i^2 + \tau_j^{2}}  =   \sum_{i = 1}^{m}  \omega_{i, j}^{(p)} \frac{(d_i - \mu^{}_j)^2}{(\sigma_i^2 + \tau_j^{2})^2 } 
 \label{eq:variance1} %\\
 \end{eqnarray}

\begin{algorithm}[tb]
\textbf{Input}: 
$\bm{d} = (d_1, \ldots, d_m)$, $\bm{\sigma^2} = (\sigma^2_1, \ldots, \sigma^2_m)$, $K$\\
\textbf{Output}: 
\vspace{-1.5em}
\begin{align*}
\bm{\theta} & = (\mu_1, \ldots, \mu_K, \tau_1, \ldots, \mu_K),  \\ 
\bm{\pi} & = (\pi_1, \ldots, \pi_K)
\end{align*}  
\vspace{-1.5em}
\begin{algorithmic}[1] %[1] enables line numbers
\STATE Let converge = False.
\STATE Initialize model parameters in previous iteration as $\bm{\theta}^{(p)}, \bm{\pi}^{(p)}$.
\WHILE{Not converge} 
\STATE Compute values of $\omega_{i,k}^{(p)}$ using Equation~\eqref{eq:e-step}, $\forall i, \forall k$.
\STATE Update $\bm{\theta}, \bm{\pi}$ based on Equations~\eqref{eq:weight1},~\eqref{eq:mu1} and ~\eqref{eq:variance1}
\STATE Compute the marginal log-likelihood of the data $log (f(\bm{d},  |\bm{\theta}, \bm{\pi} ))$ given the new parameters.
\IF {Change in log-likelihood/m is less than the specified tolerance}
\STATE converge = True
\ELSE
\STATE  $ \bm{\theta}^{(p)} = \bm{\theta}, \bm{\pi}^{(p)} = \bm{\pi}$
\ENDIF
\ENDWHILE
\STATE \textbf{return} solution.
\STATE Repeat Step 1-12 multiple times using different initial points to obtain the global optimum.
\caption{EM algorithm solving Three-Layer Heteroscedastic GMM}\label{alg: Em}
\end{algorithmic}
\end{algorithm}
The EM algorithm is summarized in Algorithm~\ref{alg: Em}. By setting $K = 3$ and $\mu_2 = 0$, the mean for positive components $\mu_1$ serves as the estimate of AES. To avoid getting trapped in stationary point\cite{wu1983convergence}, we repeat the EM algorithm multiple times with randomly initialized starting points.%~\cite{biernacki2003choosing}. 

	\textbf{Singularity issues in Gaussian mixture model. }	
When the covariance matrix is singular, the variances becomes zero, it leads to a spiky Gaussian component that "collapses" into a single point. There are several papers discussing how to address this issue. The first approach involves setting a lower bound on variance~\cite{hathaway1985constrained}. The second approach introduces a penalty term in the log-likelihood function to prevent the variance of a specific component from becoming too small~\cite{chen2008inference, ridolfi2001penalized}. In addition,~\citet{chen2008inference, chen2017consistency} introduce the conditions and provide proofs for the consistency of the MLE under finite Gaussian mixture model.
Using the second approach, the marginal log-likelihood takes the following form:
\begin{eqnarray}
\label{eq:em_penalty}
\nonumber
&&log (f(\bm{d}| \bm{\theta}, \bm{\pi} )) + log(p_m(\bm{\tau^2})) =
\sum_{i = 1}^m \sum_{k=1}^K [log(\pi_k) \\ && + log (f_k(d_i|\bm{\theta}))] 
- \frac{1}{m} \sum_{k= 1}^K( \frac{1}{\tau^2_k} + log(\tau^2_k)), 
\end{eqnarray}
where the penalty function $p_m(\bm{\tau^2})$ represents the product of K inverted Gamma distributions. Equation~\eqref{eq:variance1} can be revised to incorporate this penalty.
%(See Appendix). 

\textbf{Extensions.} This framework can be extended to any $K$ with $K > 3$, using information criteria to determine the optimal $K$. Subsequently, the estimation of AES for $K > 3$ is a weighted average of the means of these positive components, where weights are also estimated in GMM.

%Second, in this paper, we use the mean of the positive component as the estimated AES; however, it's possible to utilize the density function, such as the percentile of the fitted GMM Distribution from Algorithm~\ref{alg: Em}, for this purpose.
	
% 	so
% latently, there exists clusters
% we cannot distinguish them by eyeballing

% actually I think this is indeed a good motivation to use GMM

% e.g., in the simulaiton, by set the true lift smaller (equvalent, noiser)

% Observed ES - could be used for challenging the GMM assumption - but could be also a good motivation (the noise in the third layer is large and makes the clusters not distinguishable by eyeballing)

\section{Utility Theory: Bayesian Optimal Effect Size}
\label{section:bayesian}
	%In Section~\ref{section:fre}, 
     In the previous section, we proposed a modeling approach that focuses on estimating the mean effect size. 
	Such an approach is easy to explain to experimentation service users. 
	However, in many real-world scenarios, users prioritize identifying the most suitable parameters for achieving maximum gains rather than just maximizing the accuracy of estimation. 
	Motivated by the real needs, we propose a utility-maximization framework as our second approach.

	The SSD problem is essentially a trade-off between information gain and decision accuracy. 
	We define an utility (reward) function that considers all the related components, with the goal of maximizing the overall gain from experimentation. 
% 	which uses AES as a parameter, with the goal of identifying the optimal AES which makes trade-off between the cost of conducting experiment and making precise launch decisions. 
	The utility include  the following components: 
\begin{enumerate}
    \item Weekly experimentation cost $c$: we assume there is a fixed and pre-specified \textit{weekly cost for running the experiment}, which may include the \textit{personnel and hardware} to support this specific experiment to collecting more samples, 
    the \textit{opportunity cost for the experimentation service}, 
    and the  \textit{opportunity cost for the experimenters}. 
    \item The \textit{customers impact on the treatment group during the experiment}.
     For example, if the treatment has clear negative impacts on customers, running it longer will result in greater losses for company . 
    \item The \textit{customers impact from making the launch recommendation} on weeks $t$. This term concerns the decision accuracy, i.e., we would like to launch those new features that indeed have positive customer impacts. 
\end{enumerate}

We demonstrate this approach with the problem of finding an one-size-fits-all default effect size. 
We assume there are $m$ historical experiments. 
Our goal is to learn a default effect size that works best for all experiments on average, and we estimate the expectation using the empirical average over past experiments. 
The objective of finding the appropriate AES becomes maximizing the following utility function:
%	Therefore, we proposes the following utility function, composed of the three aforementioned components. The objective of finding the appropriate AES becomes maximizing the utility function, as following:	
\begin{align}
 \label{eqn:objective}
 \nonumber
&\argmax_{\mu > 0} \;\;
    \sum_{i=1}^m
    \Mean_{\delta_i \sim \N(d'_i, \sigma_i^2)}
    \Bigg[
    \underbrace{-c_i \cdot (T_i(\mu)-1)
    }_\text{Opportunity Cost}\\
     \nonumber
    &+ 
    \underbrace{
    \big( \delta_i \!\cdot\! N_{i, 1:T_i(\mu), T}
    \big)
    }_{\mathclap{\substack{\text{Impact during the experiment}\\ \text {(relative to control)}}}} 
    \nonumber
    + 
    \underbrace{
      u_2(\delta_i, H, T_i(\mu))\I( \pi(\bm{Y}_{T_i(\mu), i}) = 1)
    }_\text{Launch impact (relative to control)}
    \Bigg]\\
     \nonumber
    &= \argmax_{\mu > 0} \sum_{i=1}^m
    \Bigg[
    \underbrace{-c_i \cdot (T_i(\mu)-1)
    }_\text{Opportunity Cost} \\
     \nonumber
    &+  
    \underbrace{
    \big( d'_i \!\cdot\! N_{i, 1:T_i(\mu), T}
    \big)
    }_{\mathclap{\substack{\text{Impact during the experiment}\\ \text {(relative to control)}}}} 
    +  
    \underbrace{
      u_2(d'_i, H, T_i(\mu))\I( \pi(\bm{Y}_{T_i(\mu), i}) = 1)
    }_\text{Launch impact (relative to control)}
    \Bigg] \\
\end{align}
where the notations are as follows: 
\begin{itemize}
\item $\delta_i$ is the per-customer per-week treatment effect for experiment $i$. 
\item $d'_i$ is the posterior mean of the effect size at the end of the experiment and $\sigma_i^2$ is the posterior variance
\item  The subscript $i=1,...,m$ indexes the past similar experiments.
\item $T_i(\mu)$ is the recommended duration week for experiment $i$, given AES $\mu$. In Frequentist hypothesis testing, $T(\mu)$ is the minimum number of weeks required for its $power(\mu, \sigma_., n_., \alpha)$ in Equation~\eqref{eq:power} to exceed a predetermined threshold, such as 80\%. 
\item $N_{i, 1:t, g}$ denotes the total number of customers in group $g \in \{T , C \}$ observed up to week $t$ for experiment $i$.
\item $c_i$ is the weekly opportunity cost for experiment $i$. $c_i$ is proportional to the total sample sizes of each experiments.
\item $u_2 (\delta, H, t)$ represents the impact of launching the treatment feature on the whole population at week $t$, given a pre-specified time horizon $H$. Typically, we study one year impact, i.e. $H = 52$ weeks. One example of estimating one year launch impact is: $u_2 (\delta, H, t) = \delta * (H - t) * (\sum_{g \in \{T, C\}}N_{i, 1:t, g})$
\item $\pi$ is the decision rule. $\pi(.) = 1$ indicates the experiment is launched, 0 otherwise. 
\item $\bm{Y}_{t, i}$ contains all observed outcomes from customers triggered up to week $t$ for experiment $i$.
\end{itemize}
The equality in Equation~\eqref{eqn:objective} is due to the linearity of expectation when $u_2$ is a linear function in the ATE (which is true in our case).

\textbf{Optimization}. Equation~\eqref{eqn:objective} is a one-dimensional optimization problem, for which we can efficiently find a good solution. 
We use grid search in our prototype.

% Hence, we propose the following framework and the reward function, drawing inspiration from the high level idea in~\cite{wan2023experimentation}. 

% \textbf{Computation}. Unlike sequential decision making defined in~\cite{wan2023experimentation}, Equation~\ref{eqn:objective} is a one-dimensional optimization problem, hence it can be solved efficiently without the need of Reinforcement learning algorithm. One can use either \textit{scipy.optimize} or grid research. 

\textbf{Opportunity cost. } One challenge to this approach is that how to select the opportunity cost c. This value is provided by business team or estimated through analysis of past launch experiments. Choosing a larger c will result in a greater AES, thus a shorter duration, as the longer experiments incurs higher costs. The impact of $c$ has been studied in Figure 4 in~\citet{wan2023experimentation}. 
% \textbf{Discussion and Extension. } This method can be implemented and extended as follows. 
% \begin{enumerate}
%     % \item 
% \end{enumerate}

\section{Experiments}
We have proposed two approaches to select AES given past similar experiments. 
% To train Three-Layer GMM in Section~\ref{section:fre}, we only need the observed effect size from past similar experiments. Utility theory method described in Section~\ref{section:bayesian} requires the observed effect size, observed sample size and variance to compute the recommended duration week.
In this section, we compare the performance of both approaches against baseline methods.

\subsection{Accuracy Comparison with Simulation}
\label{section:simulations}

In this section, we compare the accuracy of different AES estimators. 
Since the ground truth is unknown in real data, we use simulation for this study. 
% Given the necessary to know the ground truth effect size, we use simulated data. 
As the utility-based approach requires more information of experiments and its primary objective differs from the other approaches, we postpone its analysis to the next section.

\begin{figure}[t]
  \centering
   \includegraphics[width=0.7\columnwidth]{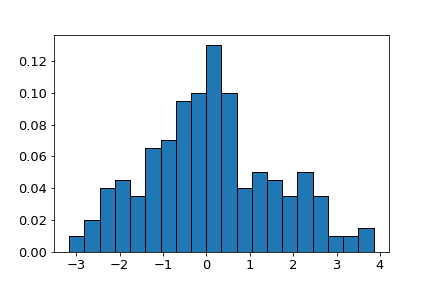}
  \caption{Histogram of simulated observed effect size $d_i$. }
    \label{fig:section6_hist}
\end{figure}

\textbf{Dataset.} We simulate $observed$ effect size from three-layer heteroscedastic GMM model in~\eqref{eq:three-layer-model}, with $K = 3$, $(\mu_1, \mu_2, \mu_3) = (2, 0, -2)$, $(\tau_1, \tau_2, \tau_3) = (0.5, 0.5, 0.5)$, $(\pi_1, \pi_2, \pi_3) = (0.2, 0.6, 0.2)$, $\sigma_i^2 \sim \textit{Inverse-Gamma}(3, 0.7)$, $m = 200$. The histogram of simulated $d_i$ is shown in Figure~\ref{fig:section6_hist}. Due to the presence of heteroscedasticity $\sigma_i^2$ alongside the variance in each Gaussian component $\tau_k^2$, it is challenging to distinguish individual components through visual inspection alone. 

\textbf{Metrics.} To quantify the accuracy of AES estimators, we repeat the simulations $iter = 50$ times and use the Mean Square Error (MSE) and Mean Absolute Error (MAE):
\begin{eqnarray*}
MSE &=& \sum_{i= 1}^{iter} \frac{(Estimation_i - Actual_i)^2}{iter}, \label{eq:MSE} \\
MAE &=& \sum_{i= 1}^{iter} \left|\frac{Estimation_i - Actual_i}{iter}\right|.
\label{eq:MAE}
\end{eqnarray*}

\begin{table}[b]
	\centering
	\small
	\begin{tabular}{|@{\,}c@{\,}|c | c |}
		\hline
		  & MSE & MAE \\
		\hline
		Pooled-MLE &  0.709 & 0.842 \\
		\hline
		Two-layer GMM & 0.017 & 0.180\\
		\hline	
		Three-layer GMM& 0.003  &  0.137\\
		\hline
	\end{tabular}
    \caption[Optional captions]{Accuracy Comparison: MSE and MAE comparison among pooled effect size, standard two-layer GMM and the proposed three-layer heteroscedastic GMM.}
    	\label{table:forecast_performance}
\end{table}

\begin{figure}[t]
  \centering
    \includegraphics[width=0.9\columnwidth]{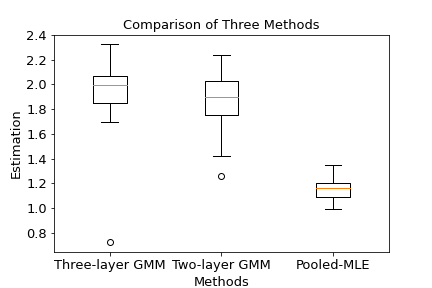}
  \caption{Boxplots comparing the AES estimations among pooled effect size, standard two-layer GMM and the proposed three-layer heterocasdestic GMM. Ground truth is 2. }
    \label{fig:section6_box}
\end{figure}

\textbf{Methods and Results.} Table~\ref{table:forecast_performance} compares the MSE and MAE for the pooled effect size (Pooled-MLE), standard two-layer GMM (Two-layer GMM) and the proposed three-layer  heteroscedastic GMM (Three-layer GMM). For Pooled-MLE, the MLE of parameter $\mu_0$ is computed using the model in Equation~\eqref{eq:base_line} on all positive observed effect sizes $d_i$. Two-layer GMM uses Gaussian Mixture in \textit{sklearn} package~\cite{scikit-learn} with $K = 3$. Three-layer GMM, the proposed method, uses the proposed EM-Algorithm~\ref{alg: Em} with the penalty term~\eqref{eq:em_penalty}, $K = 3$ and $\mu_2 = 0$. Each simulation uses $10$ different starting points to avoid local optimum (One of these starting points is initialized using the k-means clustering). Both Two-layer GMM and Three-layer GMM uses the estimated mean for the positive components as AES and $tolerance = 10^{-3}$. Table~\ref{table:forecast_performance} shows that Three-layer GMM performs slightly better than Two-layer GMM. The p-value from a two-sample t-test is 0.036, indicating the Three-layer GMM has significantly better accuracy.
As expected, Pooled-MLE is underestimated. GMM outperforms Pooled-MLE notably in cases where we lack information about whether the experiment's effect is a true positive or not. The Boxplot in Figure~\ref{fig:section6_box} illustrates a similar conclusion.

\begin{table*}[t]
       \centering
       \setlength{\tabcolsep}{0.5em} % for the horizontal padding
        % \hspace{-1cm}
		\renewcommand{\arraystretch}{1}
				\begin{tabular}{lllllllll}
			\hline
			&  \makecell{ Estimated \\ AES ($\%$)  } 
			& \makecell{(Empirical) \\ False \\ Positive}
			& \makecell{(Empirical) \\ False \\ Negative } 
			& \makecell{ Avg\\Weeks } 
			& \makecell{ Avg\\Opportunity\\Cost (D) } 
			& \makecell{ Avg\\Launch\\Impact (D)} 
			& \makecell{ Avg\\Impact\\During \\Exper (D)} 
			& \makecell{ \textbf{Avg}\\\textbf{Reward } \\(D)} 
			\\
\hline
%Pooled\_MLE&0.19\%&0.2\%&0.2\%&3.72&0.5&2.38&0.0&1.88\\
Pooled-MLE&0.05\%&0.0\%&0.0\%&3.99&1.0&2.61&0.01&1.61\\

Two-layer GMM&0.22\%&0.29\%&0.29\%&3.63&0.41&2.42&-0.0&2.01\\

Three-layer GMM&0.15\%&0.13\%&0.11\%&3.83&0.65&2.55&0.01&1.91\\

Utility-maximization&1.58\%&1.17\%&0.99\%&2.3&0.02&2.22&0.0&\textbf{2.2}\\

\hline
\end{tabular}
		\caption{Meta-analysis results. Recall that all utility-related metrics
share the same unit D, the meaning of which is omitted due to confidentiality.}
		\label{tab:fre_ops}
\end{table*}

\subsection{Meta-Analysis with Real Experiments}
In the previous section, we evaluated the estimation accuracy of different methods. However, the accuracy is just one facet to consider. 
Recall that the choice of the AES is always a trade-off between experimentation cost and accuracy. Therefore, it is hard to define a single optimal solution. 
In this section, we compare the two proposed approaches with baseline methods on real-world experimental data in terms of their empirical performance over a few metrics. 

\textbf{Dataset and setup.} 
We collect a dataset of 3,300 historical experiments conducted in the past two years within Amazon, each having a duration of 4 weeks. %Due to the data privacy policies, we refrain from sharing those data.
$d_i$ represents the observed standardized effect size derived from a specific standardization formulation used in the company at week 4. 
Given the absence of the ground-truth effect size, we adopt a heuristic yet easy-to-explain approach that uses the observed effect size at the end of the 4 weeks as the empirical ground-truth effect size. 
%We've included a simulation study in Appendix D for reproducibility purposes, since we cannot share this real dataset due to confidentiality constraints. 

We set the maximum duration as $4$ weeks for the analysis. 
For each estimated AES obtained from different methods, we plug in them, along with the observed sample size and sample variance (or their predicted versions as described  in~\citet{richardson2022bayesian}), into power Equation~\eqref{eq:power} to calculate the statistical power at week $i= 1, 2, 3, 4$. 
We define the recommended duration as the minimum number of weeks ($\leq 4$) required to attain a power of 80\%. 
In Frequentist setting, we use the one-sided two-group Welch's t-test~\cite{welch1947generalization} as decision policy $\pi(.)$. This implies that the decision policy $\pi(.) = 1 $ for launch if the p-value is less than the significance level of $0.05$  and the ATE is positive.

\begin{table*}
       \centering
        % \hspace{-2cm}
        \setlength{\tabcolsep}{0.4em} % for the horizontal padding
		\renewcommand{\arraystretch}{1}
		\begin{tabular}{lllllllll}
			\hline
			&  \makecell{ Estimated \\ AES  } 
			& \makecell{(Empirical)\\False \\  Positive}
			& \makecell{ (Empirical)\\False \\ Negative } 
			& \makecell{ Avg\\Weeks } 
			& \makecell{ Avg\\Opportunity\\Cost ($10^4$) } 
			& \makecell{ Avg\\Launch\\Impact ($10^4$)} 
			& \makecell{ Avg\\Impact\\During \\Exper ($10^4$)} 
			& \makecell{ \textbf{Avg}\\\textbf{Reward } \\($10^4$)} 
			\\
\hline
Pooled-MLE&0.782\%&0.13\%&0.1\%&4.0&2.8&5.88&0.0&3.08\\

Two-Layer GMM&1.041\%&0.13\%&0.27\%&3.89&2.53&5.83&0.01&3.31\\

Three-layer GMM&\textbf{1.005}\%&0.13\%&0.17\%&3.92&2.6&5.87&0.01&3.28\\

Utility-maximization&1.100\%&0.13\%&0.37\%&3.82&2.41&5.8&0.01&\textbf{3.41}\\

\hline
\end{tabular}
		\caption{Simulated data analysis results (Ground truth is 1).}
		\label{tab:fre_appendix}
\end{table*}

% Subsequently, we compare the decision changes made at the recommended duration with those reached over the entire four-week period, which we treated as the ground truth.

\textbf{Methods.} We compare the proposed estimated AESs from Three-Layer GMM and utility theory-based optimal effect size (Utility-maximization) with those from two-layer GMM and Pooled-MLE. 
%In Section~\ref{section:simulations} pooled-MLE performs poorly when estimated AES using experiments with positive effect sizes. To rectify this, we enrich the information provided to this method by training the MLE using significant positive experiments (i.e. $d_i >0$ and $p-value < \alpha$). 
The settings for Two-layer GMM, Three-layer GMM and Pooled\_MLE are the same as 
the previous section.
%Section~\ref{section:simulations}.  
Utility-maximization uses grid-search to find the optimal effect size within the set $\{0.02\%, 0.04\%, \ldots, 2\%\}$.

\textbf{Metrics.} 
We consider the following metrics:
\begin{itemize}
  \item[1.] The percentage of empirical false positives (proportion of incorrectly
detecting significantly positive effect when the true effect is flat/negative) and the percentage of empirical false negatives (proportion of incorrectly
detecting flat/negative effect when the true effect is indeed positive). These two metrics reflect the (empirical) Type-I and Type-II errors.
  \item[2.] The average utility in Equation~\eqref{eqn:objective} and its three components, including the opportunity cost, the impact during the experiment, and the launch impact.
  % \item[3.] , which includes the three components mentioned above, is our main objective.
\end{itemize}

\textbf{Results.} 
We present results from 3300 experiments in Table~\ref{tab:fre_ops}. The pooled MLE has the smallest decision errors. However, it comes at the expense of requiring longer experiment duration, leading to higher costs.
The proposed Utility-maximization method outperforms other methods and generates the highest average cumulative reward, with a desired balance between the cost of experimentation and making  correct launch decisions. 
%The trend on how the selected AES influences the utility function is in Appendix C.

\textbf{More simulations.} Due to confidentiality constraints, the above dataset cannot be shared. To ensure reproducibility, we conducted a simulation study with 3000 simulated experiment trajectories at weeks 1, \ldots 4: (1) Simulate weekly sample size following beta-geometric distribution~\cite{richardson2022bayesian}. The beta distribution parameters, $\alpha$ and $\beta$, are drawn from uniform distributions: $\alpha \sim Uniform(0.1, 1)$, $\beta \sim Uniform(4, 60)$. Each arm assumes a total of 10K customers. (2) Sample the observed effect size $d_{i, t}$ for experiment $i$ at week $t$ from the three-layer GMM model~\eqref{eq:three-layer-model} with three Gaussian components $(\mu_1, \mu_2, \mu_3) = (-1, 0, 1)$, $(\tau_1, \tau_2, \tau_3) = (0.3, 0.5, 0.3)$, $(\pi_1, \pi_2, \pi_3) = (0.2, 0.6, 0.2)$. Given $\sigma^2_g = 500$, $\sigma_{i}^2$ are computed using sample size and Equation~\eqref{eq:mean_diff}. (3) Total weekly opportunity cost is $c = 4 \cdot 10^6$, decomposed to each experiment according to their final weeks’ sample sizes. 
The observed effect size on last week $\{d_{i, 4}\}_{i = 1}^m$ are used to fit MLE and GMM methods.
Utility-maximization uses grid-search to find the optimal effect size
within the set \{0.1, 0.2, \ldots, 5\}. Results in Table~\ref{tab:fre_appendix} demonstrate that the Three-layer GMM yields the most precise estimation (ground truth is 1), whereas the utility-based approach achieves the highest utility.

\section{Conclusion}
AES places an central role in duration recommendation, yet little attention has been drawn to its specification, particularly for large online experimentation services. 
In this paper, we propose two approaches to estimate AES from a large number of historical online experiments. 
The first approach introduces a novel three-layer GMM to account for experiment heteroscedasticity.  
The second approach finds the optimal AES that maximizes the expected utility. 
Through simulations and a large-scale meta-analysis using real experiments from Amazon, we conclude that
the first proposed ensures a high estimation accuracy and the second proposed approach leads to a significant gain in the expected utility. With the provided flexibility,  one can choose either of these approaches to achieve their specific goal.

For the GMM-based method, exploring other hierarchical models and leveraging experiment-specific features to recommend personalized effect size are important next steps. 
Bayesian non-parametric models \citep{orbanz2010bayesian} or structures that have been explored in bandits \citep{wan2021metadata} are good starting points. 
For the utility-based method, we can also easily extend it to provide personalized effect size (and hence personalized duration) recommendation, by replacing the end-of-horizon posteriors of the $m$ experiments with the posterior for the target experiment at the duration recommendation time point. 
Besides, exploring the performance with other ATE estimators such as covariate adjusted estimators \citep{masoero2023leveraging} is also an interesting next step. 

% for both methods, we use two-sample t-test in our study; it is of interest to study the performance with other options. 
% For example, this paper primarily centers around frequentist power, while we can leverage 
% the distribution learned via GMM in Bayesian hypothesis testing, or by defining $T(\mu)$  as Bayesian utility theory (or Hybrid Bayesian-frequenstist power~\cite{du2016bayesian}) and $\pi(.)$ as Bayesian decision rules in the utility-based method. 
% The interaction of these methods with covariate adjusted estimators \citep{masoero2023leveraging} is also an interesting next step. 

% \item , where  is defined upon frequentist power. However, it is easy to extend this framework into Bayesian Hypothesis testing by define $T(\mu)$ using 
% \item This approach can be 

\bibliography{aaai24.bib}

\end{document}